\documentclass[letterpaper, 10 pt, conference]{ieeeconf}

\IEEEoverridecommandlockouts                              % This command is only needed if 
\usepackage[utf8]{inputenc}
\usepackage{comment}
\usepackage{booktabs}

\usepackage{graphicx}
\usepackage{caption}
\usepackage{subcaption}
\usepackage{amsmath}
\usepackage{verbatim}
\usepackage{amsfonts}
\usepackage{multirow}
\usepackage{changepage}
\usepackage{tabularx}
\usepackage{adjustbox}
\usepackage{mathtools}
\usepackage{amssymb}
\usepackage{soul}
\usepackage{color}
\usepackage[noend]{algpseudocode}

\usepackage{color}
\usepackage{tikz}
\usepackage{flushend}
\usetikzlibrary{shapes.geometric, arrows}
\usepackage[linesnumbered,ruled]{algorithm2e}
\usepackage{multirow}
\usepackage{soul,color}
\usepackage{graphicx}
\usepackage[colorlinks]{hyperref}
\usepackage{fancyhdr}
\usepackage{comment}
\usepackage{epsfig}
\usepackage{bm}
\usepackage{float}
\usepackage{cuted}
\usepackage{tcolorbox}
\usepackage{xcolor}
\usepackage{mdframed}
\usepackage{listings}
\allowdisplaybreaks

\makeatletter
\def\algbackskip{\hskip-\ALG@thistlm}
\makeatother

\newcommand\oprocendsymbol{\hbox{$\square$}}
\newcommand\oprocend{\relax\ifmmode\else\unskip\hfill\fi\oprocendsymbol}

\usepackage{cite}

\newcommand\red[1]{{\color{red} #1}}
\newcommand\blue[1]{{\color{blue} #1}}

\newcommand\orange[1]{{\color{orange} #1}}

\newtheorem{remark}{Remark}

\renewcommand{\theenumi}{(\roman{enumi}}
\renewcommand{\baselinestretch}{0.97}

\mdfdefinestyle{custombox}{%
    backgroundcolor=gray!10, % Define a light gray background
    linewidth=1pt,           % Border thickness
    linecolor=black,         % Border color
    innertopmargin=10pt,     % Top margin inside the box
    innerbottommargin=10pt,  % Bottom margin inside the box
    innerrightmargin=10pt,   % Right margin inside the box
    innerleftmargin=10pt     % Left margin inside the box
}

\makeatletter
\let\NAT@parse\undefined
\makeatother

\begin{document}
%In case you encounter the following error:
%Error 1010 The PDF file may be corrupt (unable to open PDF file) OR
%Error 1000 An error occurred while parsing a contents stream. Unable to analyze the PDF file.
%This is a known problem with pdfLaTeX conversion filter. The file cannot be opened with acrobat reader
%Please use one of the alternatives below to circumvent this error by uncommenting one or the other
%\pdfobjcompresslevel=0
%\pdfminorversion=4

% See the \addtolength command later in the file to balance the column lengths
% on the last page of the document

% The following packages can be found on http:\\www.ctan.org
%\usepackage{graphics} % for pdf, bitmapped graphics files
%\usepackage{epsfig} % for postscript graphics files
%\usepackage{mathptmx} % assumes new font selection scheme installed
%\usepackage{times} % assumes new font selection scheme installed
%\usepackage{amsmath} % assumes amsmath package installed
%\usepackage{amssymb}  % assumes amsmath package installed

\title{\LARGE \bf
Graph-Grounded LLMs: Leveraging Graphical Function Calling to Minimize LLM Hallucinations
}

\author{
Piyush Gupta\textsuperscript{1*} \hspace{1.2cm}  Sangjae Bae\textsuperscript{1} \hspace{1.2cm} 
David Isele\textsuperscript{1} 
\thanks{
\textsuperscript{1} Honda Research Institute, San Jose, CA, 95134, USA.}
\thanks{
\textsuperscript{*} Corresponding author \texttt{\{piyush\_gupta\}@honda-ri.com } \
}
\thanks{This work has been supported by Honda Research Institute, USA.}
}

% \author{Piyush Gupta$^{1}$,
%         Sangjae Bae$^{1}$%
% % \thanks{$^{1}$Honda Research Institute, USA, Inc. San Jose, CA 95134, USA
% %         {\tt\small \{alexandre\_mirandaanon, sbae, manish\_saroya, disele\}@honda-ri.com}}%
% }

\maketitle
% \thispagestyle{empty}
% \pagestyle{empty}

%%%%%%%%%%%%%%%%%%%%%%%%%%%%%%%%%%%%%%%%%%%%%%%%%%%%%%%%%%%%%%%%%%%%%%%%%%%%%%%%
\begin{abstract}

The adoption of Large Language Models (LLMs) is rapidly expanding across various tasks that involve inherent graphical structures.  Graphs are integral to a wide range of applications, including motion planning for autonomous vehicles, social networks, scene understanding, and knowledge graphs. Many problems, even those not initially perceived as graph-based, can be effectively addressed through graph theory. However, when applied to these tasks, LLMs often encounter challenges, such as hallucinations and mathematical inaccuracies. To overcome these limitations, we propose Graph-Grounded LLMs, a system that improves LLM performance on graph-related tasks by integrating a graph library through function calls. By grounding LLMs in this manner, we demonstrate significant reductions in hallucinations and improved mathematical accuracy in solving graph-based problems, as evidenced by the performance on the NLGraph benchmark. Finally, we showcase a disaster rescue application where the Graph-Grounded LLM acts as a decision-support system.
\end{abstract}

%===================================================
% INTRODUCTION
%===================================================

\section{Introduction}
\label{intro}

In recent years, Large Language Models (LLMs) have gained significant traction and found applications across a broad spectrum of tasks that extend well beyond traditional text completion. These applications include multi-robot task planning~\cite{kannan2024smart}, autonomous driving~\cite{ gupta2023interaction, gupta2024towards, mao2023gpt, gupta2022towards}, human-robot interactions~\cite{wang2024lami, gupta2021robust, tanneberg2024help, gupta2024structural}, structured commonsense reasoning~\cite{kwon2024toward, kang2025gflowvlm}, and knowledge probing~\cite{zheng2023kglens}. Despite their impressive performance in tasks that require common sense reasoning, LLMs continue to face challenges such as hallucinations and mathematical inaccuracies. These challenges hinder their deployment in real-world applications, particularly those that involve safety-critical systems.

Parallel to these developments, researchers are exploring the use of LLMs for tasks involving implicit graph structures\cite{jin2024large, ren2024survey, liu2024can}. Graphs are foundational in engineering, as they simplify complex problems through visual representations and the principles of graph theory, serve as memory-efficient data structures, and benefit from optimized algorithms that enhance problem-solving capabilities. Robust graph reasoning and efficient operations are essential in diverse applications, from knowledge and social networks to motion planning. Empowering LLMs to reliably and accurately tackle graph-related challenges could unlock their full potential in addressing engineering and robotics problems, ultimately paving the way for their integration into safety-critical systems.

In~\cite{huang2022language}, the researchers examine how LLMs can convert high-level instructions into concrete actions within structured synthetic environments. In~\cite{ammanabrolu2021learning}, the authors develop methods to continuously update and maintain structured world representations as states evolve, thereby facilitating operations in interactive, real-world settings. For structured commonsense reasoning, LLMs are provided with natural language inputs and tasked with producing graphs—such as event graphs~\cite{tandon2019wiqa}, reasoning graphs~\cite{madaan2021could}, or argument explanation graphs~\cite{saha2021explagraphs}—that encapsulate the underlying relationships. In knowledge probing, LLMs uncover implicit connections and pathways among a broad network of entities and concepts~\cite{creswell2022selection}. Furthermore,\cite{chen2024exploring} investigates the potential of LLMs in performing graph node classification. In~\cite{hu2024scalable}, authors leverages swarm intelligence to solve graph reasoning
problems by assigning an LLM agent to each
node, allowing it to focus on processing its own information and communicating with neighbors.    
Lastly, in multi-robot mission planning, \cite{gupta2025generalized} utilized LLMs to construct hierarchical tree representations of missions, which then aid in devising individual plans for each robot. Collectively, these studies underscore the growing trend of employing LLMs for tasks and environments characterized by implicit graphs and structured representations.

Despite advances that have improved LLM performance in graphical tasks, several critical questions remain about their ability to effectively handle graph-related problems. Can LLMs explicitly interpret textual descriptions of graphs, map them onto grounded conceptual spaces, and execute structured operations on these representations? Early research in this domain~\cite{wang2024can} indicates that while LLMs demonstrate preliminary graph reasoning abilities, their performance deteriorates as problem complexity increases. Additionally, they are prone to brittle failures when confronted with spurious correlations. Furthermore, the advantages of prompting and in-context learning diminish as graph-based challenges become more intricate.

Given these limitations, a hybrid approach is warranted. LLMs should be utilized for high-level reasoning, while computationally intensive mathematical operations are best handled by traditional software, particularly with the availability of efficient graph libraries. This strategy reduces the computational burden on LLMs, enhancing both reliability and accuracy in solving graph-related problems.

To tackle these challenges, we propose Graph-Grounded LLMs, a novel system that combines a graph library with the function-calling capabilities of LLMs to reduce hallucinations in graph-related tasks. The function-calling feature is increasingly gaining traction for leveraging large-scale APIs to enhance LLM accuracy across various domains~\cite{du2024anytool, ocker2024tulip}. 
%\bae{Any existing approaches that relate graphs and LLM? such as https://arxiv.org/pdf/2410.05130} 
Unlike previous approaches, our method employs multiple rounds of closed-loop function calling, enabling the LLM to iteratively construct graphs and execute graphical operations to logically and robustly solve implicit graph problems with high accuracy across diverse domains.

To rigorously evaluate our approach, we employ the Natural Language Graph (NLGraph) benchmark~\cite{wang2024can}, a comprehensive evaluation benchmark designed for graph-based problem solving in natural language. The standard NLGraph benchmark consists of 6022 problems comprising eight distinct graph reasoning tasks, each varying in complexity based on both the size of the graph network and the difficulty of the problem itself. These tasks range from fundamental concepts, such as determining connectivity and detecting cycles, to more advanced challenges like computing maximum flow and solving bipartite matching.

Our work makes three key contributions to the existing literature. First, we present Graph-Grounded LLMs, a system that integrates a graph library with the function-calling capabilities of LLMs to minimize hallucinations in graph-related tasks. 
Second, we conduct a rigorous evaluation using the NLGraph benchmark, demonstrating near 100\% accuracy on most tasks while analyzing failure cases to highlight potential limitations and areas for improvement. Third, we show a disaster response application where the Graph-Grounded LLM acts as a natural language decision-support system, gathering environmental data, mapping the scenario to a graph problem, assisting in
response planning, answering scenario-related queries, and facilitating dynamic re-planning.

The remainder of the paper is organized as follows. In Section~\ref{sec:graph_grounded_LLMs}, we introduce Graph-Grounded LLMs along with motivating examples. Section~\ref{Sec: NLGraph} details the NLGraph benchmark and the various graph-related tasks used for evaluation. In Section~\ref{sec:results}, we present the evaluation results on the NLGraph benchmark and analyze the failure cases. In Section~\ref{sec:application}, we present a disaster response application where the Graph-Grounded LLM acts as a decision-support system in natural language. Section~\ref{sec:Limitations} discusses the limitations of our approach, and finally, Section~\ref{sec:conclusions} concludes the paper.

\section{Graph-Grounded LLMs}\label{sec:graph_grounded_LLMs}

Recent advances in LLMs enhanced their performance in tasks with implicit graph structures. However, despite the preliminary graph reasoning ability of the LLMs, they suffer from hallucinations (fact alteration), mathematical inaccuracies, and non-repeatability of solutions, often limiting their usage in real-world applications. To this end, we present Graph-Grounded LLMs, a system that leverages the function-calling ability of the LLMs with a user-built graph library to reduce the computational overhead of the LLMs in graph-related tasks. The following are the advantages offered by the Graph-Grounded LLMs:

\begin{itemize}
    \item \textbf{Leveraging Pre-existing Libraries:} Efficient graph libraries such as $\mathrm{networkx}$ enable easy development of user-defined functions, allowing use of optimized algorithms and software for improved performance.

    \item \textbf{Task-Agnostic Graph Library:} A user-built library can be designed as a general-purpose toolset, independent of specific tasks. This allows LLMs to flexibly utilize a broad range of graph operations to solve diverse problems.

    \item \textbf{Reduced LLM Overhead:} Since the graph computations are handled externally, the LLM does not need to track all context and intermediate steps. This minimizes reliance on memorization, enhances accuracy, and reduces computational burden.

    \item \textbf{Scalability to Larger Graphs:} While LLMs struggle with increasing problem size, traditional graph-processing libraries efficiently handle large graphs. By offloading computations, Graph-Grounded LLMs maintain high accuracy even for complex, large-scale tasks.

    \item \textbf{Robust Edge-Case Handling:} User-defined functions enable explicit handling of both inherent problem edge cases and those introduced by LLMs, enhancing the reliability of Graph-Grounded LLMs. For instance, if an LLM mistakenly generates a directed graph instead of an undirected one, a function requiring an undirected graph can internally convert it, ensuring correctness without requiring LLM correction.

    \item \textbf{Error-Guided Self-Correction:} LLMs are prone to hallucinations, often leading to incorrect responses. By incorporating structured error handling in user-defined functions, LLMs can self-reflect and refine their outputs. For instance, if a bipartite graph function returns an error on a non-bipartite graph, the LLM can correct its approach by regenerating a valid bipartite graph.
    
\end{itemize}

\begin{figure}%[ht]
    \centering
    \includegraphics[width=\columnwidth]{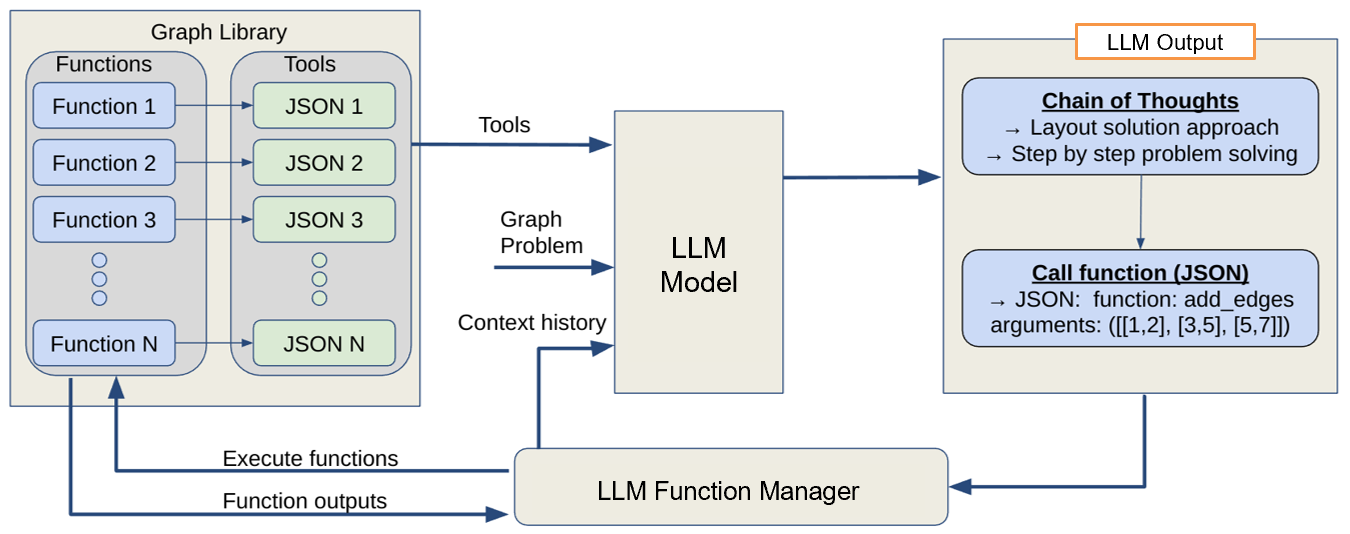}
    \caption{Schematic of the Graph-Grounded LLM. A graph library containing various graph functions is created, with each function described in JSON format and provided to the LLM as tools alongside the context and graph problem description. The LLM employs chain-of-thought reasoning to devise a solution strategy and determine the necessary function calls. These function calls are then executed by the LLM function manager, which returns the computed outputs to the LLM for further reasoning and problem-solving.}
    \label{fig:Framework}
\end{figure}

Fig.~\ref{fig:Framework} presents a schematic of the Graph-Grounded LLM. The framework includes a user-built graph library containing various graph functions, each accompanied by a JSON-formatted description. These JSON descriptions are provided to the LLM as tools, along with the context and user-specified graph problem. This setup enables the LLM to comprehend the available graph functions and their operations.

The graph library is designed to address both inherent and LLM-induced edge cases. We employ a 0-chain-of-thought reasoning~\cite{wei2022chain, yu2023towards} approach, where the LLM is prompted to think step-by-step through the problem, invoking graph functions as needed via function calls. Once a solution strategy is formulated, the LLM determines the required function calls, which are executed by the function manager. The computed outputs are then returned to the LLM for further reasoning and problem-solving.

In this approach, a graph is explicitly constructed, and all requested operations are performed systematically, ensuring accuracy. The process involves multiple rounds of closed-loop function requests and executions before the LLM arrives at the final solution. Additionally, built-in error checking and handling within the graph library enable the LLM to self-reflect and self-correct its graph construction and function requests in case of mistakes.

\subsection{Graph Library}
We design the graph library with 29 commonly used graph functions, enabling our system to handle a wide range of graph-related tasks. The library is task-agnostic, featuring a general-purpose toolset for graph construction and operations. Key functions include graph construction (directed/undirected, weighted/unweighted), adding nodes and edges, computing shortest distances and paths, topological sorting, maximum flow computation, and identifying connected components, among others. Internally, these functions leverage the efficient $\mathrm{networkx}$ library for implementation. Additionally, the library is built with robust error handling and edge-case considerations, allowing the LLM to self-reflect and self-correct its mistakes, improving reliability and performance. 
%\bae{I know you have already narrated the key functions, but can you add a snapshot of graph function definitions? and/or general strategies of defining them -- it might be possible that just few functions are called in most cases while other functions are barely or never called. If that's the case, any strategies for further refining them? What I am trying to mainly say: this graph library is the key contribution, so adding details would be helpful -- it should be worth more than a 1/4 page.}

\begin{figure}%[ht]
    \centering
    \includegraphics[width=0.7\columnwidth]{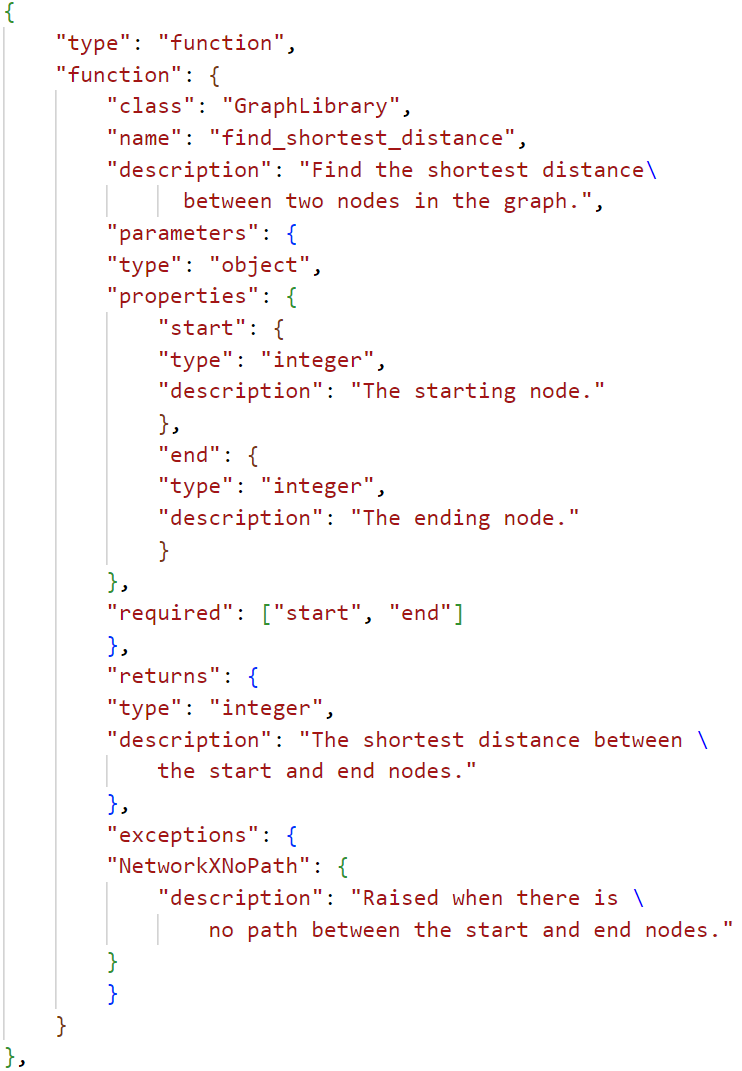}
    \caption{JSON description of the find\_shortest\_distance function for LLM.}
    \label{fig:JSON}
\end{figure}

For each function in the graph library, we create a JSON description that includes details such as the function's purpose, its arguments, and their respective data types. This structured representation helps the LLM understand the function's utility and ensures it generates arguments with the correct data types. Fig.~\ref{fig:JSON} shows an example JSON description of the $\mathrm{find\_shortest\_path}$ function. Based on our experience, simple function descriptions are sufficient for the LLM to grasp the function's utility. 

% \begin{lstlisting}
% {
%     "type": "function",
%     "function": {
%         "class": "GraphLibrary",
%         "name": "find_shortest_distance",
%         "description": "Find the shortest distance between two nodes in the graph.",
%         "parameters": {
%         "type": "object",
%         "properties": {
%             "start": {
%             "type": "integer",
%             "description": "The starting node."
%             },
%             "end": {
%             "type": "integer",
%             "description": "The ending node."
%             }
%         },
%         "required": ["start", "end"]
%         },
%         "returns": {
%         "type": "integer",
%         "description": "The shortest distance between the start and end nodes."
%         },
%         "exceptions": {
%         "NetworkXNoPath": {
%             "description": "Raised when there is no path between the start and end nodes."
%         }
%         }
%     }
% }
% \end{lstlisting}
\subsection{Motivating Examples}

We now present two motivating examples using the $\mathrm{gpt-3.5-turbo-0125}$ model, highlighting cases where the standalone LLM produces incorrect answers, while Graph-Grounded LLMs effectively utilize function calls to construct the relevant graph, execute appropriate graph operations, and logically derive the correct solution.

\begin{mdframed}[style=custombox,  frametitle={\blue{Social Network} \ \ 
\includegraphics[width=0.3\columnwidth]{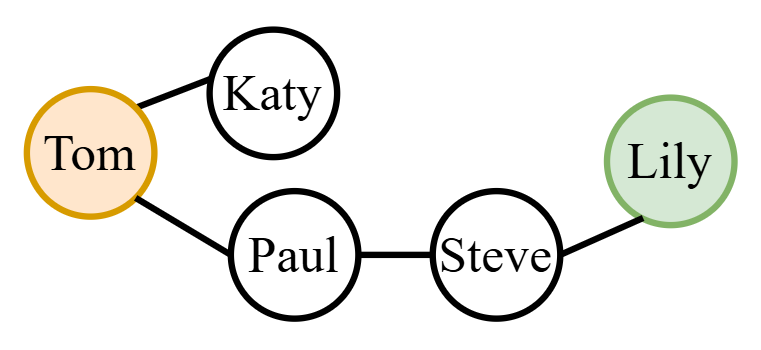}   
    }]

\footnotesize
\noindent
\textbf{Stand-Alone LLM:}

\red{System:} Ask for clarification if a user request is ambiguous. First think step by step and then you can find the solution.

\noindent\rule{\textwidth}{1pt}

\blue{User:} Consider a bunch of friends. Tom is a friend of Katy and Paul, Paul is a friend of Steve, and Steve is a friend of Lily. Only friends can pass messages to each other. How many message passing would it take if Tom wants to send a message to Lily? Think step by step. What should be the procedure to do that?

\noindent\rule{\textwidth}{1pt}
\red{assistant:} 
To find out how many message passing it would take for Tom to send a message to Lily, we need to determine the shortest path of friends between them. Here's the step-by-step procedure:

\begin{enumerate}
    \item Tom wants to send a message to Lily.
    \item Tom is friends with Katy and Paul, so Tom can directly pass the message to Katy and Paul.
    \item We know that Paul is a friend of Steve.
    \item Steve is a friend of Lily.
    \item Therefore, the shortest path for Tom to send a message to Lily is: Tom $\rightarrow$ Katy $\rightarrow$ Paul $\rightarrow$ Steve $\rightarrow$ Lily.

\end{enumerate}
To summarize:
\begin{itemize}
    \item Tom passes the message to Katy (1 message passing).
    \item Katy passes the message to Paul (2 message passing).
    \item Paul passes the message to Steve (3 message passing).
    \item Steve passes the message to Lily (4 message passing).
\end{itemize}

It would take 4 message passings for Tom to send a message to Lily. The problem has been solved. \red{[Incorrect]}

\noindent\rule{\textwidth}{2pt}

\noindent
\textbf{Graph-Grounded LLM:}

\red{System:} You can use the given functions through tool\_calls to arrive at a solution. After each tool\_call, the tool will provide you with the output of the function. Don't make assumptions about what values to plug into functions. Ask for clarification if a user request is ambiguous. Before making any function calls, first think step by step, determine a procedure to solve the problem, and then you can make tool\_calls to find the solution. Once the problem is solved, provide the final answer.

\noindent\rule{\textwidth}{1pt}

\blue{User:} [Same as above]

\noindent\rule{\textwidth}{1pt}
\red{assistant:} To determine how many message passings it would take for Tom to send a message to Lily, we need to find the shortest path between Tom and Lily in the friend network described. Here is the step-by-step procedure to find the shortest path:

\begin{enumerate}
    \item Create a graph representing the friend network with the given relationships.
    \item Add nodes for Tom, Katy, Paul, Steve, and Lily.
    \item Add edges to connect the friends as per the relationships: - Tom is a friend of Katy and Paul. - Paul is a friend of Steve.  - Steve is a friend of Lily.
    \item Find the shortest path between Tom and Lily in the friend network.
\end{enumerate}

\red{assistant:}  
%[\{`id': `call\_Y71gXKwPEjhSVfz3al2FD8Sh', `type': `function', 
`\orange{function}': \{`name': `GraphLibrary\_init', `arguments': `\{``directed": true, ``weighted": false\}'\}
%\}]

\noindent\rule{\textwidth}{1pt}
\begin{flushright}
\blue{function:}  GraphLibrary constructor was called
%with `arguments': `\{``directed": true, ``weighted": false\}'
\end{flushright}
\noindent\rule{\textwidth}{1pt}
\red{assistant:}  
%[\{`id': `call\_Q7IYNCTeq2FWsH6blMXm8zbo', `type': `function', 
`\orange{function}': \{`name': `add\_nodes', `arguments': `\{ ``nodes": [0, 1, 2, 3, 4] \}'\}
%\}]

\noindent\rule{\textwidth}{1pt}
\begin{flushright}
\blue{function:} add\_nodes was called 
%with arguments \{ ``nodes": [0, 1, 2, 3, 4]\} 
\end{flushright}
\noindent\rule{\textwidth}{1pt}
\red{assistant:}  
%[\{`id': `call\_GNFbxr4j17o1wC6SXoacjDnM', `type': `function', 
`\orange{function}': \{`name': `add\_edges', `arguments': `\{ ``edges": [[0, 1], [0, 2], [1, 3], [3, 4]] \}'\}
%\}]
\noindent\rule{\textwidth}{1pt}
\begin{flushright}
\blue{function:} add\_edges was called
%with arguments \{ ``edges": [[0, 1], [0, 2], [1, 3], [3, 4]]\} 
\end{flushright}
\noindent\rule{\textwidth}{1pt}
\red{assistant:}  
%[\{`id': `call\_O2Kl4bMWCg4vruud3mEVv3hq', `type': `function', `
\orange{function}': \{`name': `find\_shortest\_path', `arguments': `\{`start": 0, ``end": 4\}'\}
%\}]

\noindent\rule{\textwidth}{1pt}
\begin{flushright}
\blue{function:}  find\_shortest\_path was called 
%with arguments \{``start": 0, ``end": 4\} 
and resulted in [0, 1, 3, 4]
\end{flushright}

\noindent\rule{\textwidth}{1pt}
\red{assistant:}
The shortest path between Tom (node 0) and Lily (node 4) is: Tom $\rightarrow$ Katy $\rightarrow$ Steve $\rightarrow$ Lily.\\
Therefore, it would take 3 message passings for Tom to send a message to Lily through Katy and Steve. The problem is solved. \blue{[Correct]}
\end{mdframed}

\begin{mdframed}[style=custombox,  frametitle={\blue{Disconnected graph}  \ \ 
\includegraphics[width=0.3\columnwidth]{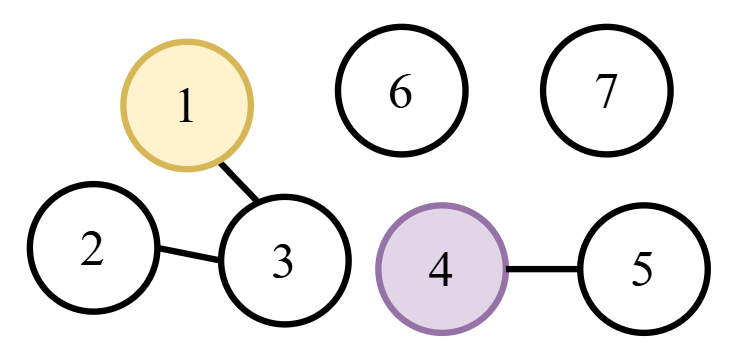}   
    } ]
\footnotesize
\noindent
\textbf{Stand-Alone LLM:}

\red{System:} [Same as previous example]

\noindent\rule{\textwidth}{1pt}

\blue{User:} In a graph with 7 nodes and following edges: [1 3], [2,3], [4,5], find the distance between node 1 and node 4. Think step by step. What should be the procedure to do that?

\noindent\rule{\textwidth}{1pt}
\red{assistant:} 
To find the distance between node 1 and node 4 in the given graph, you can follow these steps:

\begin{enumerate}
    \item Visualize the graph: Draw a representation of the graph with 7 nodes labeled 1 to 7. Connect nodes 1 and 3, nodes 2 and 3, as well as nodes 4 and 5 with edges.

    \item Find the shortest path: Determine the shortest path between node 1 and node 4. Since there is no direct edge connecting them, you need to find the route that involves the fewest number of edges.

    \item Analyze the connections: Examine the connections in the graph to identify the possible paths. From node 1, you can reach node 3 directly or go through node 2 to reach node 3. From node 3, you can reach node 4 directly or go through node 5 to reach node 4.

    \item Calculate the distance: Compare the two possible paths: 1 $\rightarrow$ 3 $\rightarrow$ 4 and 1 $\rightarrow$ 3 $\rightarrow$ 5 $\rightarrow$ 4. Count the number of edges in each path to determine the distance between node 1 and node 4. In this case, the first path has two edges, while the second path has three edges.
\end{enumerate}

Therefore, the distance between node 1 and node 4 in the given graph is 2 edges. \red{[Incorrect]}

\noindent\rule{\textwidth}{2pt}

\noindent
\textbf{Graph-Grounded LLM:}

\red{System:} [Same as previous example]

\noindent\rule{\textwidth}{1pt}

\blue{User:} [Same as above]

\noindent\rule{\textwidth}{1pt}
\red{assistant:} To find the distance between node 1 and node 4 in a graph with 7 nodes and the given edges [1,3], [2,3], and [4,5], we can follow these steps:

\begin{enumerate}
    \item Initialize a graph instance.

    \item Add 7 nodes to the graph.

    \item Add edges [1,3], [2,3], and [4,5] to the graph.

    \item Find the shortest path between node 1 and node 4.

    \item Calculate the distance of the shortest path.
    
    \item Return the distance between node 1 and node 4.

\end{enumerate}

Let's proceed with these steps and use the provided functions to perform the necessary operations.
\begin{flushleft}
    \red{assistant:}  
%[\{`id': `call\_EX90ToLJgcPSV58j3oMNcNcx', `type': `function', 
`\orange{function}': \{`name': `GraphLibrary\_init', `arguments': `\{\}'\} \ldots [Omitted for space]
%\}]
\end{flushleft}
% \noindent\rule{\textwidth}{1pt}
% \begin{flushright}
% \ \ \vdots \ \ \ [Omitted for space]
% \end{flushright}
% \noindent\rule{\textwidth}{1pt}
% \begin{flushright}
% \blue{function:} GraphLibrary constructor called
% \end{flushright}
% \noindent\rule{\textwidth}{1pt}
% \red{assistant:}  
% %[\{`id': `call\_6PMgdKbwauSjgdNKQs9fo2y4', `type': `function', `
% \orange{function}': \{`name': `add\_nodes', `arguments': `\{ ``nodes": [1, 2, 3, 4, 5, 6, 7] \}'\}
% %\}]
% \noindent\rule{\textwidth}{1pt}
% \begin{flushright}
% \blue{function:} add\_nodes was called with arguments \{ ``nodes": [1, 2, 3, 4, 5, 6, 7]\} 
% \end{flushright}
% \noindent\rule{\textwidth}{1pt}
% \red{assistant:} 
% %[\{`id': `call\_765N0QGqDxmB1QHpHIu9dvJs', `type': `function', `
% \orange{function}': \{`name': `add\_edges', `arguments': `\{ ``edges": [[1, 3], [2, 3], [4, 5]] \}'\}
% %\}]
% \noindent\rule{\textwidth}{1pt}
% \begin{flushright}
% \blue{function:} add\_edges was called with arguments \{ ``edges": [[1, 3], [2, 3], [4, 5]]\} 
% \end{flushright}
% \noindent\rule{\textwidth}{1pt}
% \red{assistant:}  
% %[\{`id': `call\_Yav9gJPaO3cvMuSOCoQqDn94', `type': `function', `
% \orange{function}': \{`name': `find\_shortest\_path', `arguments': `\{ ``start": 1, ``end": 4 \}'\}
% %\}]
% \noindent\rule{\textwidth}{1pt}
% \begin{flushright}
% \blue{function:} find\_shortest\_path was called with arguments \{ ``start": 1, ``end": 4 \} and resulted in No path between 1 and 4 
% \end{flushright}
\noindent\rule{\textwidth}{1pt}
\red{assistant:}
I apologize, but there is no path between node 1 and node 4 according to the given edges. Therefore, the distance between node 1 and node 4 is undefined. \blue{[Correct]} 
\end{mdframed}

These motivating examples illustrate the advantages of Graph-Grounded LLMs over stand-alone LLMs for graph-related tasks.

%===================================================
% METHODS
%===================================================

\section{NLGraph Benchmark}\label{Sec: NLGraph}

\begin{figure*}[ht]
    \centering
    \includegraphics[width=0.8\linewidth]{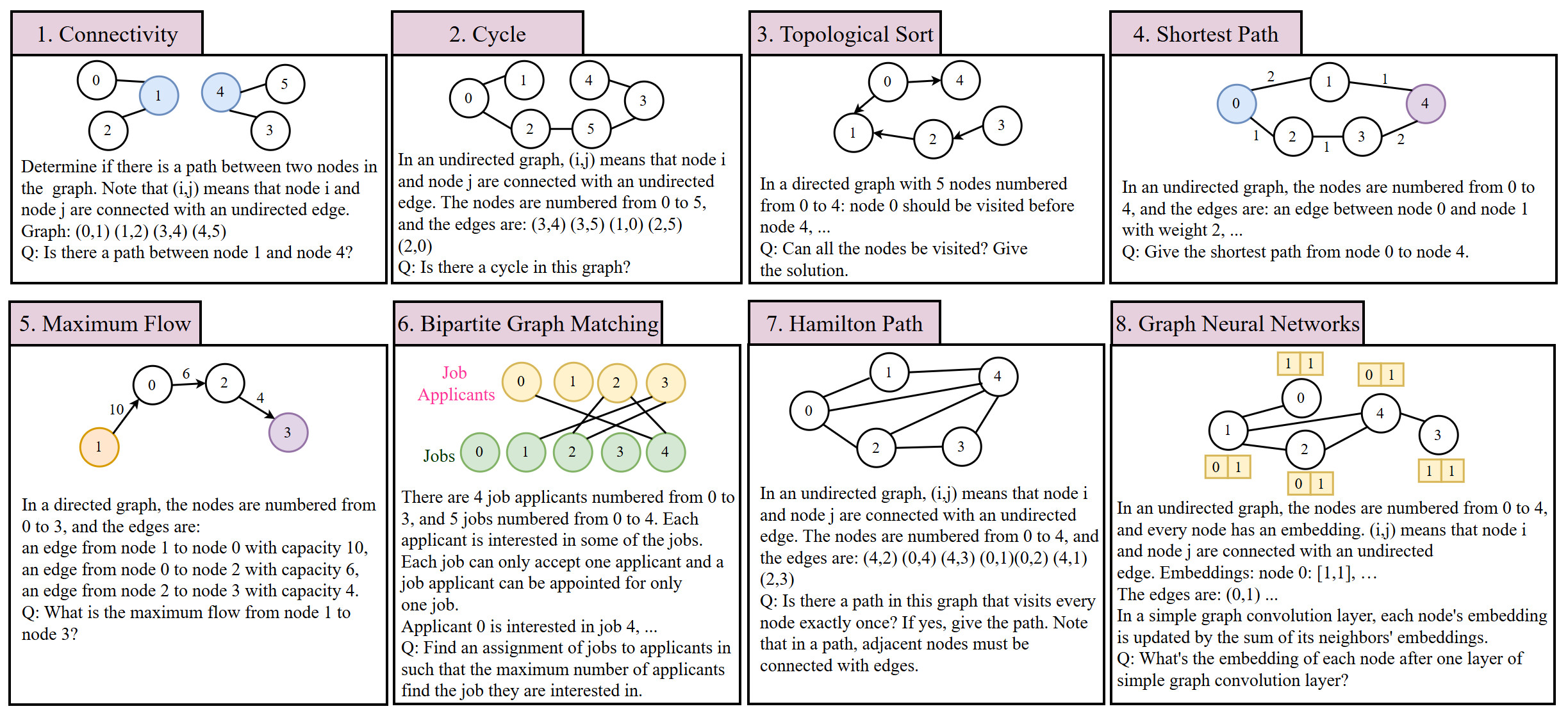}
    \caption{Overview of NLGraph Benchmark Tasks: This illustration presents eight graph-related tasks of varying complexity, each accompanied by sample natural language prompts used with LLMs. The image is recreated from~\cite{wang2024can}.}
    \label{fig:NLGraph}
\end{figure*}

In this work, we evaluate the performance of Graph-Grounded LLM on the NLGraph benchmark~\cite{wang2024can}, which comprises 6,022 problems spanning eight distinct graph reasoning tasks. The dataset features problems of varying difficulty, with factors such as graph size, network sparsity, and numeric range influencing complexity. Consequently, some tasks are divided into easy, medium, and hard subsets, while others are categorized into easy and hard levels. The eight task categories in the benchmark include:

\begin{enumerate}
\item Connectivity: In this task, the LLM determines whether two given nodes $u \in \mathcal{V}$ and $v \in \mathcal{V}$ in an undirected graph $\mathcal{G} = \{\mathcal{V}, \mathcal{E}\}$ are connected.

\item Cycle: In this task, the LLM determines whether there exists a cycle in an undirected graph $\mathcal{G} = \{\mathcal{V}, \mathcal{E}\}$.

\item Topological Sort: In this task, the LLM is required to produce a valid topological ordering based on a set of node order constraints. Since multiple correct solutions exist, an external program is used to verify the validity of the ordering generated by the LLM.

\item Shortest Path: In this task, the LLM determines the shortest path between two given nodes $u \in \mathcal{V}$ and $v \in \mathcal{V}$ in an undirected graph $\mathcal{G} = \{\mathcal{V}, \mathcal{E}\}$ along with the path length.

\item Maximum Flow: In this task, the LLM generates a routing plan to maximize the flow from the source node $s \in \mathcal{V}$ to the sink node $t \in \mathcal{V}$ in a graph $\mathcal{G} = \{\mathcal{V}, \mathcal{E}\}$. Each edge $e \in \mathcal{E}$ is associated with a capacity $c$, representing the maximum flow it can carry.

\item Bipartite Graph Matching: In this task, the LLM is provided with a list of job applicants, each interested in a subset of available jobs, and is tasked with determining a maximum matching between the applicants and the jobs. In graph theory, this equates to finding a maximum matching in a bipartite graph, identifying a set of edges with no common nodes that maximizes the total number of matches. An external program is employed to verify both the validity and optimality of the LLM's solution.

\item Hamilton Path: In this task, the LLM is tasked with finding a valid Hamilton path in an undirected graph $\mathcal{G} = \{\mathcal{V}, \mathcal{E}\}$. A Hamilton path is one that visits every node exactly once, and an external program is used to verify the correctness of the LLM's solution.

\item Graph Neural Network (GNN): In this task, given an undirected graph 
$\mathcal{G} = \{\mathcal{V}, \mathcal{E}\}$ and a two-dimensional embedding 
$x_i$ for each node, the LLM performs $\ell$ layers of message passing. At each layer, a node's embedding is updated by summing the embeddings of all its neighboring nodes. 

\end{enumerate}

An overview of the NLGraph benchmark tasks is shown in Fig.~\ref{fig:NLGraph}. The illustration presents the eight graph-related tasks of varying complexity, each accompanied by sample natural language prompts used with LLMs.

\section{Results}\label{sec:results}

\begin{table*}[]
\centering

%\begin{adjustbox}{width=1\columnwidth}
\begin{tabular}{|c|c|c|c|c|c|c|}
\hline
\textbf{Task} &
  \textbf{Difficulty} &
  \textbf{\begin{tabular}[c]{@{}c@{}}Number \\ of Tasks\end{tabular}} &
  \textbf{\begin{tabular}[c]{@{}c@{}}Stand-Alone LLM \\ (Correct)\end{tabular}} & \textbf{\begin{tabular}[c]{@{}c@{}}Stand-Alone LLM \\ (Average Time in s)\end{tabular}} & 
  \textbf{\begin{tabular}[c]{@{}c@{}}Graph-Grounded \\ LLM (Correct)\end{tabular}} &  \textbf{\begin{tabular}[c]{@{}c@{}}Graph-Grounded LLM \\ (Average Time in s)\end{tabular}} \\ \hline
  Connectivity &
  Easy &
  352 &
  346 (98.30\%) & 2.89 &
  \textbf{352 (100.00\%)} & 11.68 \\ \hline
Connectivity &
  Medium &
  1200 &
  1131 (94.25\%) & 3.30 &
  \textbf{1200 (100.00\%)}& 11.81\\ \hline
Connectivity &
  Hard &
  680 &
  628 (92.35\%) & 4.61 &
  \textbf{680 (100.00\%)}& 18.06\\ \hline
Cycle &
  Easy &
  150 &
  89 (59.33\%) & 7.16 &
  \textbf{149 (99.33\%)} & 12.39\\ \hline
Cycle &
  Medium &
  600 &
  347 (57.83\%) & 12.10 &
  \textbf{600 (100.00\%)} & 14.01\\ \hline
Cycle &
  Hard &
  400 &
  206 (51.50\%) & 13.28 &
  \textbf{400 (100.00\%)} & 19.62\\ \hline
Topological sort &
  Easy &
  180 &
  42 (23.33\%) & 7.96 &
  \textbf{143 (79.44\%)} & 14.76\\ \hline
Topological sort &
  Medium &
  450 &
  4 (0.89\%) & 14.69 &
  \textbf{343 (76.22\%)} & 21.05\\ \hline
Topological sort &
  Hard &
  180 &
  0 (0.00\%) & 10.54 &
  \textbf{145 (80.56\%)} & 43.17\\ \hline
Shortest Path &
  Easy &
  180 &
  96 (53.33\%) & 6.21 &
  \textbf{180 (100.00\%)} & 13.52\\ \hline
Shortest Path &
  Hard &
  200 &
  48 (24.00\%) & 10.55 &
  \textbf{200 (100.00\%)} &  16.01\\ \hline
Maximum Flow &
  Easy &
  150 &
  \begin{tabular}[c]{@{}c@{}}25  (16.66\%)\end{tabular} & 7.84 &
  \textbf{\begin{tabular}[c]{@{}c@{}}149  (99.33\%)\end{tabular}} & 13.06\\ \hline
Maximum Flow &
  Hard &
  200 &
  \begin{tabular}[c]{@{}c@{}}10  (5.00\%)\end{tabular} & 11.39 &
  \textbf{\begin{tabular}[c]{@{}c@{}}196  (98.00\%)\end{tabular}} & 23.79\\ \hline
Bipartite Matching &
  Easy &
  300 &
  219 (73.00\%) & 6.22 &
  \textbf{297 (99.00\%)} & 28.32\\ \hline
Bipartite Matching &
  Hard &
  210 &
  121 (58.00\%) & 17.73 & 
  \textbf{204 (97.00\%)} & 45.67 \\ \hline
Hamilton Path &
  Easy &
  150 &
  48 (32.00\%) & 7.44 & 
  \textbf{150 (100.00\%)} & 15.15\\ \hline
Hamilton Path &
  Hard &
  200 &
  17 (8.50\%) & 11.44 &
  \textbf{200 (100.00\%)} & 17.01 \\ \hline
GNN &
  Easy &
  100 &
  81 (81.00\%) & 11.43&
  \textbf{100 (100.00\%)} & 16.85\\ \hline
GNN &
  Hard &
  140 &
  61 (44\%) & 22.04 & 
  \textbf{140 (100\%)} & 21.46 \\ \hline
\end{tabular}
%\end{adjustbox}
\caption{Performance on the NLGraph benchmark}
\label{tab:NLgraph}
\end{table*}

\begin{table*}[]
\centering
%\begin{adjustbox}{width=1\columnwidth}
\begin{tabular}{|c|c|c|c|c|c|c|}
\hline
\textbf{Task} &
  \textbf{Difficulty} &
  \textbf{\begin{tabular}[c]{@{}c@{}}Number \\ of Tasks\end{tabular}} &
  \textbf{\begin{tabular}[c]{@{}c@{}}Stand-Alone LLM \\ (Correct)\end{tabular}} & \textbf{\begin{tabular}[c]{@{}c@{}}Stand-Alone LLM \\ (Average Time in s)\end{tabular}} & 
  \textbf{\begin{tabular}[c]{@{}c@{}}Graph-Grounded \\ LLM (Correct)\end{tabular}} &  \textbf{\begin{tabular}[c]{@{}c@{}}Graph-Grounded LLM \\ (Average Time in s)\end{tabular}} \\ \hline
Topological sort & Easy   & 180 & 87 (48.33\%) & 5.39 & \textbf{179 (99.44\%)} & 15.36\\ \hline
Topological sort & Medium & 450 & 24 (5.33\%) & 17.20 & \textbf{450 (100\%)} & 22.06  \\ \hline
Topological sort & Hard   & 180 & 0 (0.00\%) & 10.50  & \textbf{158 (87.78\%)} & 40.85 \\ \hline
\end{tabular}
%\end{adjustbox}
\caption{Improved performance in Topological Sort with the modified problem prompt.}
\label{tab:topological_sort}
\end{table*}

We now present the evaluation results on the eight different graph reasoning tasks in the NLGraph benchmark (Section~\ref{Sec: NLGraph}). Table~\ref{tab:NLgraph} compares the performance of a stand-alone LLM with that of a Graph-Grounded LLM on these tasks. The average computation time for the Graph-Grounded LLM is generally higher than that of a stand-alone LLM due to multiple rounds of closed-loop function calls. Notably, some problems exhibit significantly longer computation times than others. To provide a more representative measure, we report average computation times after rejecting these outliers in computation time. For all evaluations, we use the $\mathrm{gpt-4-0613}$ model as the chosen LLM.

Among the tasks, connectivity and cycle detection are considered the easiest. The stand-alone LLM achieves over 90\% accuracy in the connectivity task and above 50\% across the cycle tasks. In contrast, the Graph-Grounded LLM attains nearly 100\% accuracy in both. For more complex tasks, such as shortest path, Hamilton path, and Graph Neural Network evaluation, the Graph-Grounded LLM consistently achieves 100\% accuracy, whereas the stand-alone LLM performs significantly worse. Similarly, in the maximum flow and bipartite matching tasks, the stand-alone LLM exhibits much lower accuracy, while the Graph-Grounded LLM achieves close to 100\% accuracy.

The few observed failures primarily stem from two factors: hallucinations caused by ambiguities in the problem prompt and the model exceeding its context limit (maximum context length for $\mathrm{gpt-4-0613}$ model is 8,192 tokens) when handling extremely large graphs. The most notable drop in performance for the Graph-Grounded LLM occurs in the topological sort task. This issue is attributed to the original NLGraph problem prompt, which asks whether all nodes in a graph with multiple node order constraints can be visited. This phrasing is similar to the Hamilton path problem, which seeks a path visiting each vertex exactly once. As a result, the LLM sometimes incorrectly invokes the Hamilton path function instead of the topological sort function. Since our Hamilton path implementation only applies to undirected graphs, this misinterpretation leads to failures in cases where a directed acyclic graph is provided.

To address this issue, we modified the problem prompt for topological sorting from ``Can all the nodes be visited?" to ``Can all the nodes be visited in a valid topological order that satisfies these constraints?" Table~\ref{tab:topological_sort} presents the updated performance results for this task with the revised prompt. With this adjustment, the Graph-Grounded LLM achieves nearly 100\% accuracy on easy and medium tasks and 87.78\% on hard tasks. The remaining failure cases primarily stem from exceeding the LLM's context length when processing extremely large graphs.

\begin{remark}
    The Hamilton path task belongs to the class of NP-complete problems. As a result, our backtracking implementation in the graph library is highly inefficient and requires a significant amount of time to solve.
\end{remark}

By analyzing the failure cases, we derive the following conclusions:

\begin{itemize}
    \item \textbf{Function Utilization Errors:} In a single failed cycle task of easy difficulty, the Graph-Grounded LLM devised a correct solution strategy but failed to utilize the available functions. Such rare failures can be mitigated by either restarting the task or employing self-consistency through majority voting~\cite{wang2022self}.

    \item \textbf{Misuse of Similar Functions:} When multiple similar functions exist in the graph library, the LLM may select an unintended function due to the availability of multiple solution approaches. In such cases, the model is highly sensitive to problem prompts, and clearer prompts are necessary to ensure that the correct functions are used by the LLM.

    \item \textbf{Context Length Limitations:} With a large number of available tools, extremely large problems may fail due to exceeding the LLM’s context length. Scalability techniques such as Retrieval-Augmented Generation (RAG)~\cite{gao2023retrieval} can help manage tool selection efficiently in such scenarios.

\end{itemize}

% \todo{typical size, Add column of average computation time}

\section{Disaster Response Application}\label{sec:application}

Graph-Grounded LLMs have broad applications in tasks involving implicit graph structures, such as inventory management, social networks, and multi-robot traffic management. In this work, we explore a disaster response application where a Graph-Grounded LLM serves as a natural language decision-support system. To enhance its access to critical data from robot sensors and the environment, we incorporate functions like get\_environment\_data, get\_environment\_map\_data etc.

\begin{figure}[ht]
    \centering
    \includegraphics[width=0.6\linewidth]{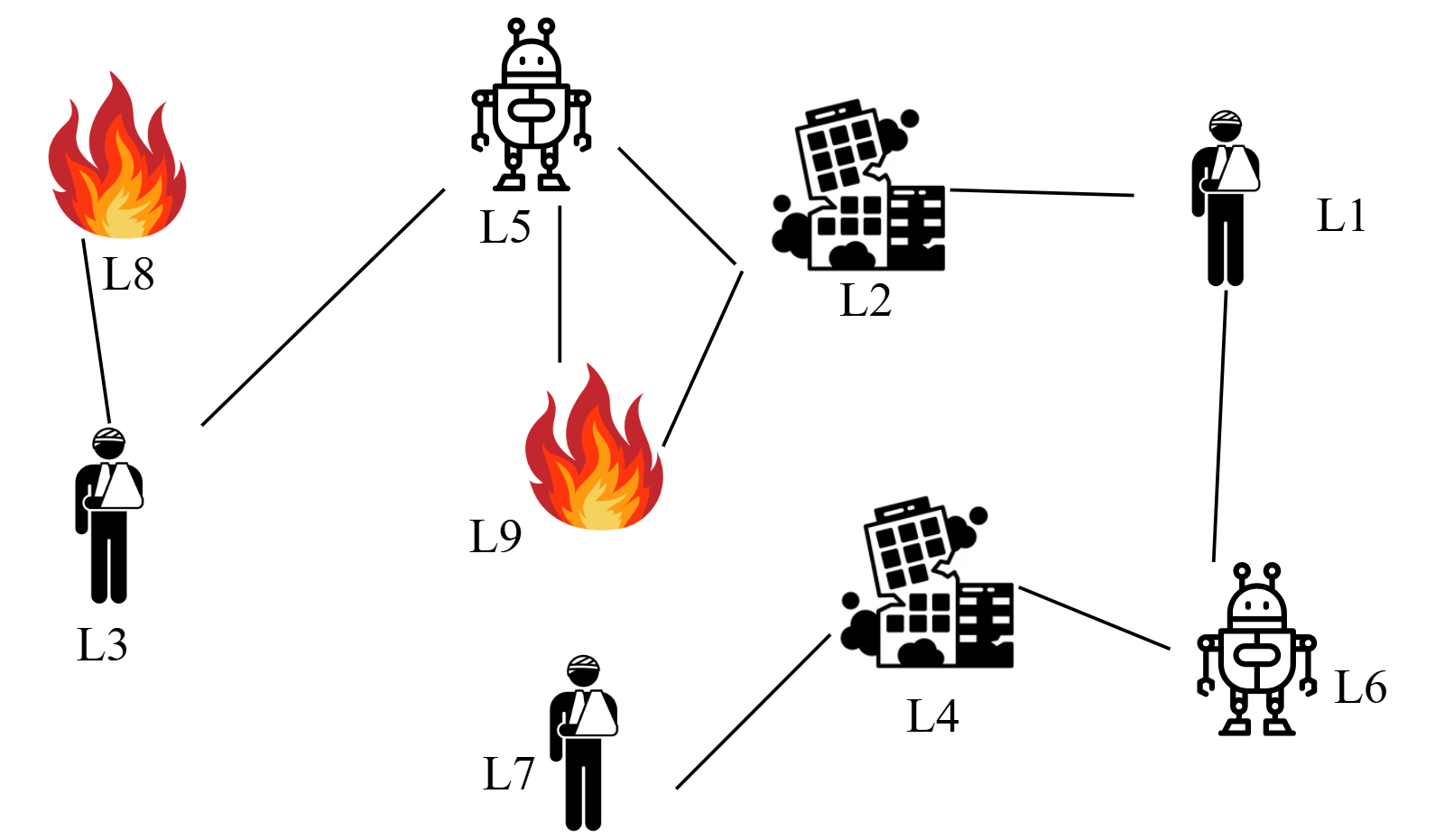}
    \caption{Illustration of the disaster response application with some fire locations, building collapse, victims and rescue robots. The Graph-Grounded LLM acts as a natural language decision support system for the disaster response mission.}
    \label{fig:search&rescue}
\end{figure}

Fig.~\ref{fig:search&rescue} illustrates the disaster response application, featuring locations with collapsed buildings, fires, victims, and rescue robots. The Graph-Grounded LLM functions as a natural language decision-support system, gathering environmental data, mapping the scenario to a graph problem, assisting in response planning, answering scenario-related queries, and facilitating dynamic re-planning.

\begin{mdframed}[style=custombox,  frametitle={\blue{Disaster \& Response Decision Support}}]
\footnotesize
\noindent
\red{System:} [Same as motivating example]

\noindent\rule{\textwidth}{1pt}

\blue{User:} Deploy rescue robots. There are survivors and fire hazards around collapsed buildings.

\noindent\rule{\textwidth}{1pt}
\red{assistant:} To deploy rescue robots, we need to understand the environment first. \textit{ \blue{We can represent the environment as a graph where nodes represent locations (like buildings, hazards, and survivor locations) and edges represent paths between these locations.}} Here are the steps we can follow:
\begin{enumerate}
 \item Retrieve the environmental data including entities, locations, and relationships.
\item Retrieve the terrain data including distances between locations.
\item Create a graph \ldots [Omitted for space] 
%based on the retrieved data. Nodes will represent locations and edges will represent paths between these locations. The weight of the edges can represent the distance or difficulty of the path.
%\item Find the shortest path from the rescue robot's location to each survivor's location, avoiding fire hazards.
%\item Deploy the rescue robots along the calculated paths.
\end{enumerate}

Let's start by retrieving the environmental data.
\begin{flushleft}
\red{assistant:}  
%[\{`id': `call\_Y71gXKwPEjhSVfz3al2FD8Sh', `type': `function', 
`\orange{function}': \{`name': `get\_environment\_data', `arguments': `\{\}'\}
\end{flushleft}
%\}]
\noindent\rule{\textwidth}{1pt}
\blue{function:} 
get\_environment\_data was called and its output is {\orange{`entities'}: [`rescue robots', `collapsed building', `survivors', `fire hazards'], \orange{`locations':} [`L1', `L2', `L3', `L4', `L5', `L6', `L7', `L8', `L9'], \orange{`relationships':} [{`type': `obstacle', `location': `L4', `hazard': `debris'}, {`type': `obstacle', `location': `L8', `hazard': `fire'}, {`type': `obstacle', `location': `L9', `hazard': `fire'}, {`type': `obstacle', `location': `L2', `hazard': `debris'}, {`type': `survivor', `location': `L3', `injury\_level': `high'}, {`type': `survivor', `location': `L7', `injury\_level': `low'}, {`type': `survivor', `location': `L1', `injury\_level': `high'}, {`type': `robot', `location': `L5', `status': `available'}, {`type': `robot', `location': `L6', `status': `available'}], \orange{`constraints':} [``Passing through debris takes 3 times longer", ``Can't go through fire"]}
\noindent\rule{\textwidth}{1pt}
% \red{assistant:}  The environmental data has been retrieved. Now, let's get the terrain data \ldots [Omitted for space] 
% %which includes distances between locations.
\begin{flushleft}
\red{assistant:}  
%[\{`id': `call\_Y71gXKwPEjhSVfz3al2FD8Sh', `type': `function', 
`\orange{function}': \{`name': `get\_environment\_map\_data', `arguments': `\{\}'\}
%\}]
\end{flushleft}
\noindent\rule{\textwidth}{1pt}
\blue{function:} 
get\_environment\_map\_data was called and its output is {\orange{`terrain':} [{`location1': `L8', `location2': `L3', `distance': 5}, {`location1': `L3', `location2': `L5', `distance': 3}, {`location1': `L5', `location2': `L2', `distance': 2}, {`location1': `L2', `location2': `L9', `distance': 4}, {`location1': `L2', `location2': `L1', `distance': 4}, {`location1': `L1', `location2': `L6', `distance': 5}, {`location1': `L6', `location2': `L4', `distance': 3}, {`location1': `L4', `location2': `L7', `distance': 4}, {`location1': `L5', `location2': `L9', `distance': 3}], \orange{`description':} [``each terrain can be traversed in both directions"]}
\noindent\rule{\textwidth}{1pt}
\red{assistant:} Now that we have both the environmental and terrain data, we can create a graph to represent the environment. \blue{\textit{The graph will be weighted (to represent distances or difficulty of paths) and undirected (since each terrain can be traversed in both directions).}} Let's initiate a graph instance.

\noindent
\red{assistant:}  
%[\{`id': `call\_Y71gXKwPEjhSVfz3al2FD8Sh', `type': `function', 
`\orange{function}': \{`name': `GraphLibrary\_init', `arguments': `\{``weighted": true, ``directed": false\}'\}

\noindent\rule{\textwidth}{1pt}
\begin{flushright}
\blue{function:}  GraphLibrary constructor called
\end{flushright}
\noindent\rule{\textwidth}{1pt}
\red{assistant:} The graph instance has been initiated. Now, let's add the locations as nodes to the graph. The locations are L1, L2, L3, L4, L5, L6, L7, L8, and L9.

\noindent
\red{assistant:}  
%[\{`id': `call\_Q7IYNCTeq2FWsH6blMXm8zbo', `type': `function', 
`\orange{function}': \{`name': `add\_nodes', `arguments': `\{ ``nodes": [0, 1, 2, 3, 4, 5, 6, 7, 8, 9] \}'\}  \ldots [Omitted for space]
%\}]

% \noindent\rule{\textwidth}{1pt}
% \begin{flushright}
% \blue{function:} add\_nodes was called
% \end{flushright}

\noindent\rule{\textwidth}{1pt}
\red{assistant:}
% The nodes representing the locations have been added \ldots [Omitted for space]
%to the graph. Now, let's add the edges to the graph. The edges represent paths between locations and their weights represent the distances between these locations. 
From the terrain data, we have the following edges and their weights:
\begin{itemize}
    \item Edge between L8 and L3 with a distance of 5
   \item \ldots [Omitted for space]
\end{itemize}

\blue{\textit{However, we need to consider the constraints from the environmental data. Passing through debris takes 3 times longer and we can't go through fire.}} The locations with debris are L4 and L2, and the locations with fire are L8 and L9. 

Let's add the edges to the graph with these considerations.

\noindent
\red{assistant:}  
%[\{`id': `call\_GNFbxr4j17o1wC6SXoacjDnM', `type': `function', 
`\orange{function}': \{`name': `add\_edges', `arguments': `\{ ``edges": [[8, 3], [3, 5], [5, 2], [2, 9], [2, 1], [1, 6], [6, 4], [4, 7], [5, 9]], ``weights" : [15, 3, 6, 12, 12, 5, 9, 4, 9] \}'\}
%\}]

\noindent\rule{\textwidth}{1pt}
\begin{flushright}
\blue{function:} add\_edges was called
\end{flushright}

\noindent\rule{\textwidth}{1pt}
\red{assistant:} The edges have been added to the graph with their respective weights considering the environmental constraints. 

\textit{ \blue{ Now, let's find the shortest path from the rescue robot's location to each survivor's location. The rescue robots are located at L5 and L6, and the survivors are located at L3, L7, and L1. \\
We will avoid the fire hazards located at L8 and L9 by not considering these nodes in our path finding algorithm.}}

Let's start by finding the shortest path from L5 to L3, L7, and L1.

\noindent
\red{assistant:}  
%[\{`id': `call\_O2Kl4bMWCg4vruud3mEVv3hq', `type': `function', `
\orange{function}': \{`name': `find\_shortest\_path', `arguments': `\{`start": 5, ``end": 3\}'\} \ldots [Omitted for space]
%\}]

% \noindent\rule{\textwidth}{1pt}

% \begin{flushright}
%     \ \ \ \ \ \ \vdots \ \ [Omitted for space]
% \end{flushright}

% \noindent\rule{\textwidth}{1pt}
% \begin{flushright}
% \blue{function:}  find\_shortest\_path called and its output is [5, 3]
% \end{flushright}

% \noindent\rule{\textwidth}{1pt}
% \red{assistant:}  
% %[\{`id': `call\_O2Kl4bMWCg4vruud3mEVv3hq', `type': `function', `
% \orange{function}': \{`name': `find\_shortest\_path', `arguments': `\{`start": 5, ``end": 7\}'\}
% %\}]

% \noindent\rule{\textwidth}{1pt}
% \begin{flushright}
% \blue{function:}  find\_shortest\_path called and its output is [5, 2, 1, 6, 4, 7]
% \end{flushright}
% \noindent\rule{\textwidth}{1pt}
% \red{assistant:}  
% %[\{`id': `call\_O2Kl4bMWCg4vruud3mEVv3hq', `type': `function', `
% \orange{function}': \{`name': `find\_shortest\_path', `arguments': `\{`start": 5, ``end": 1\}'\}
% %\}]

% \noindent\rule{\textwidth}{1pt}
% \begin{flushright}
% \blue{function:}  find\_shortest\_path called and its output is [5, 2, 1]
% \end{flushright}

\noindent\rule{\textwidth}{1pt}
\red{assistant:} The shortest paths from the rescue robot at location L5 to the survivors have been found:
\begin{itemize}
    \item To the survivor at L3: [L5, L3]
    \item To the survivor at L7: [L5, L2, L1, L6, L4, L7]
    \item To the survivor at L1: [L5, L2, L1]
\end{itemize}

Now, let's find the shortest paths from the rescue robot at location L6 to the survivors.

\noindent
\red{assistant:}  
%[\{`id': `call\_O2Kl4bMWCg4vruud3mEVv3hq', `type': `function', `
\orange{function}': \{`name': `find\_shortest\_path', `arguments': `\{`start": 6, ``end": 3\}'\} \ldots [Omitted for space]
%\}]

% \noindent\rule{\textwidth}{1pt}
% \begin{flushright}
%     \ \ \ \ \ \ \vdots \ \ [Omitted for space]
% \end{flushright}

% \noindent\rule{\textwidth}{1pt}
% \begin{flushright}
% \blue{function:}  find\_shortest\_path was called and its output is [6, 1, 2, 5, 3]
% \end{flushright}

% \noindent\rule{\textwidth}{1pt}
% \red{assistant:}  
% %[\{`id': `call\_O2Kl4bMWCg4vruud3mEVv3hq', `type': `function', `
% \orange{function}': \{`name': `find\_shortest\_path', `arguments': `\{`start": 6, ``end": 7\}'\}
% %\}]

% \noindent\rule{\textwidth}{1pt}
% \begin{flushright}
% \blue{function:}  find\_shortest\_path was called and its output is [6, 4, 7]
% \end{flushright}
% \noindent\rule{\textwidth}{1pt}
% \red{assistant:}  
% %[\{`id': `call\_O2Kl4bMWCg4vruud3mEVv3hq', `type': `function', `
% \orange{function}': \{`name': `find\_shortest\_path', `arguments': `\{`start": 6, ``end": 1\}'\}
% %\}]

% \noindent\rule{\textwidth}{1pt}
% \begin{flushright}
% \blue{function:}  find\_shortest\_path was called and its output is [6, 1]
% \end{flushright}

\noindent\rule{\textwidth}{1pt}
\red{assistant:} The shortest paths from the rescue robot at location L6 to the survivors have been found:
\begin{itemize}
    \item To the survivor at L3: [L6, L1, L2, L5, L3]
    \item To the survivor at L7: [L6, L4, L7]
    \item To the survivor at L1: [L6, L1]
\end{itemize}

Now, we can deploy the rescue robots along these calculated paths to rescue the survivors. The rescue robot at L5 should first go to L3, then to L1, and finally to L7. The rescue robot at L6 should first go to L1, then to L7, and finally to L3.

\noindent\rule{\textwidth}{1pt}
\blue{user:} What are the locations of victims with critical health condition?

\noindent\rule{\textwidth}{1pt}
\red{assistant:} From the environmental data, the survivors with a high injury level (which we can interpret as a critical health condition) are located at L3 and L1.
\noindent\rule{\textwidth}{1pt}
\blue{user:}  Prioritize saving victims in critical condition first.
\noindent\rule{\textwidth}{1pt}
\red{assistant:} Given the critical condition of the survivors at locations L3 and L1, we should prioritize rescuing them first. Let's reassess our rescue plan:

\textit{ \blue{The rescue robot at L5 is closest to the survivor at L3, so it should go there first. Then, it should go to L1. Lastly, it should go to L7.\\
The rescue robot at L6 is closest to the survivor at L1, so it should go there first. Then, it should go to L7. Lastly, it should go to L3.\\
This way, we can ensure that the survivors in critical condition are rescued first.}}

\noindent\rule{\textwidth}{1pt}
\blue{user:} Fire expanded to L2. Update the environment.

\noindent\rule{\textwidth}{1pt}
\red{assistant:}
%[\{`id': `call\_Q7IYNCTeq2FWsH6blMXm8zbo', `type': `function', 
`\orange{function}': \{`name': `delete\_node', `arguments': `\{ ``node": 2 \}'\}
%\}]

\noindent\rule{\textwidth}{1pt}
\begin{flushright}
\blue{function:} delete\_node was called
\end{flushright}
\noindent\rule{\textwidth}{1pt}
\red{assistant:}  \textit{ \blue{The node representing location L2, where the fire has expanded, has been removed from the graph. \\
Now, we need to find new paths for the rescue robots, avoiding the fire at location L2. \\
Let's find the shortest path from the rescue robot's location to each survivor's location again.}}

Starting with the rescue robot at location L5:

\noindent
\red{assistant:}  
%[\{`id': `call\_O2Kl4bMWCg4vruud3mEVv3hq', `type': `function', `
\orange{function}': \{`name': `find\_shortest\_path', `arguments': `\{`start": 5, ``end": 3\}'\} \ldots [Omitted for space]

% \noindent\rule{\textwidth}{1pt}
% \begin{flushright}
%     \ \ \ \ \ \ \vdots \ \ [Omitted for space]
% \end{flushright}

\noindent\rule{\textwidth}{1pt}
\red{assistant:} \textit{\blue{The rescue robot at location L6 can reach the survivors at L1 and L7. There is no path to the survivor at L3 due to the fire at location L2. Given the critical condition of the survivors at locations L3 and L1, we should prioritize rescuing them first. }}

The new rescue plan is:
\begin{itemize}
    \item The rescue robot at L5 should go to L3.
    \item The rescue robot at L6 should first go to L1 and then to L7.
\end{itemize}

This way, we can ensure that the survivors in critical condition are rescued first, and all available survivors are reached.

\end{mdframed}

% \section{Approach}\label{sec:a}

%\todo{RAG}

% \subsection{Overview}

% \subsection{Simulation Overview}

% \subsection{Simulation Result}

\section{Limitations}\label{sec:Limitations}

There are several limitations of the Graph-Grounded LLMs. First, they are specifically designed to enhance LLM performance in tasks with implicit graph structures. While a similar approach can be adapted for other use cases, it requires prior knowledge of the problem's high-level scope to appropriately expand the function library. Second, the system depends on user-built functions, necessitating extensive software development, edge-case handling, and debugging before deployment. Third, due to multiple rounds of closed-loop function calls and executions before arriving at a solution, computation time is significantly higher compared to a stand-alone LLM. Fourth, integrating a user-built graph library consumes part of the LLM's context window, potentially limiting performance as the library grows. Retrieval mechanisms, such as retrieval-augmented generation, become essential to dynamically select relevant functions and mitigate this issue. Finally, an extensive toolset can lead to undesired function calls, as the LLM may attempt multiple solution approaches, sometimes selecting suboptimal or unnecessary functions.

\section{Conclusions and Future Work}\label{sec:conclusions}
Advancements in Large Language Models (LLMs) have expanded their applicability to tasks involving implicit graph structures. However, LLMs often struggle with hallucinations, mathematical inaccuracies, lack of repeatability, and performance degradation as problem complexity increases. To address these challenges, we introduced Graph-Grounded LLMs, a system that integrates a graph library with the function-calling capabilities of LLMs to enhance reliability in graph-related tasks. Our approach enables LLMs to solve implicit graph problems consistently and robustly, achieving nearly 100\% accuracy across various domains in the NLGraph benchmark.

In future work, we will explore retrieval-augmented generation to dynamically select a relevant subset of tools based on the graph problem, enhancing the scalability of our approach. Additionally, we aim to investigate automated graph library modifications and function additions by the LLM, enabling it to adapt and expand its capabilities over time.

%%%%%%%%%%%%%%%%%%%%%%%%%%%%%%%%%%%%%%%%%%%%%%%%%%%%%%%%%%%%%%%%%%%%%%%%%%%%%%%%

%%%%%%%%%%%%%%%%%%%%%%%%%%%%%%%%%%%%%%%%%%%%%%%%%%%%%%%%%%%%%%%%%%%%%%%%%%%%%%%%

%\newpage
% \vspace*{2\baselineskip} % <<<<<<<<<<< added

% \bibliographystyle{ieee}
%\clearpage

\bibliographystyle{unsrt}
\bibliography{references.bib}

\end{document}